\UseRawInputEncoding
\documentclass[conference]{IEEEtran}
\usepackage{amsmath,amsfonts}
\usepackage{algorithmic}
\usepackage{algorithm}
\usepackage{array}
\usepackage[caption=false,font=normalsize,labelfont=sf,textfont=sf]{subfig}
\usepackage{textcomp}
\usepackage{stfloats}
\usepackage{url}
\usepackage{verbatim}
\usepackage{graphicx}
\usepackage{cite}

\usepackage{hyperref}
\usepackage{url} 
\usepackage{amsmath}
\usepackage{amssymb}
\usepackage{mathtools}
\usepackage{amsthm}
\usepackage{booktabs} 

\theoremstyle{plain}
\newtheorem{theorem}{Theorem}[section]

\theoremstyle{definition}
\newtheorem{definition}[theorem]{Definition}

\theoremstyle{remark}

\usepackage{bm} 
\usepackage{multirow} 
\usepackage{multicol}
\usepackage{xcolor} 
\usepackage{textcomp} 
\usepackage{diagbox}
\usepackage[table]{xcolor} 
\definecolor{lightgray}{gray}{0.9}
\definecolor{lightblue}{rgb}{0.85,0.92,1.0}
\definecolor{lightred}{rgb}{1.0, 0.85, 0.85}

\begin{document}

\title{Which Hyperparameters Matter? \\ A Game-Theoretic Framework for Interpretable Hyperparameter Sensitivity Analysis}


\author{\IEEEauthorblockN{1\textsuperscript{st} Nyi Nyi Aung}
\IEEEauthorblockA{\textit{Department of Mechanical and Industrial Engineering} \\
\textit{Louisiana State University}\\
Baton Rouge, USA \\
naung1@lsu.edu}
\and
\IEEEauthorblockN{2\textsuperscript{nd} Heepeom Shin}
\IEEEauthorblockA{\textit{Department of Electrical and Computer Engineering} \\
\textit{Louisiana State University}\\
Baton Rouge, USA \\
hshin4@lsu.edu}
\and
\IEEEauthorblockN{3\textsuperscript{rd} Abigail Lawlor}
\IEEEauthorblockA{\textit{Department of Mechanical and Industrial Engineering} \\
\textit{Louisiana State University}\\
Baton Rouge, USA \\
alawlo2@lsu.edu}
\and
\IEEEauthorblockN{4\textsuperscript{th} Adrian Stein}
\IEEEauthorblockA{\textit{Department of Mechanical and Industrial Engineering} \\
\textit{Louisiana State University}\\
Baton Rouge, USA \\
astein@lsu.edu}
}

\maketitle

\begin{abstract}
This work presents a game-theoretic framework for interpretable hyperparameter--objective interaction analysis rather than proposing a new optimization algorithm. In the proposed framework, Shapley Effects are employed for global sensitivity analysis, while Pareto front sets are utilized to identify effective hyperparameter configurations and support early-stage model evaluation. The resulting analysis reveals which players (hyperparameters) are most influential with respect to different objectives in a given game (application). Consequently, the proposed framework provides interpretable insights into objective-aware hyperparameter interactions, enabling practitioners to guide subsequent optimization, reduce the search space, and perform early-stage model evaluation. The effectiveness of the proposed framework is demonstrated using three distinct neural network architectures across different problem domains under multi-objective settings.
\end{abstract}


\section{INTRODUCTION}
\label{subsec:introduction}
Modern machine learning models rely critically on hyperparameters that govern architecture, optimization dynamics, regularization, and inductive bias. These choices are especially consequential for deep learning systems, where training costs are high and performance varies significantly across architectures. Consequently, practitioners invest substantial computational resources into hyperparameter optimization (HPO), commonly using black-box strategies such as random search, Bayesian optimization, evolutionary methods, and bandit-based approaches \cite{hazan_hyperparameter_2018,khan_comparative_2025,kavzoglu_advanced_2022,mu_shrinkhpo_2024}. While effective, these methods provide limited insight into \emph{which} hyperparameters are truly influential for a given network class, often leading to redundant searches and excessive training cost \cite{theodorakopoulos_hyperparameter_2024,g_m_hyperparameter_2023}.

This opacity is increasingly problematic in safety-critical, resource-constrained, and human-in-the-loop settings, where interpretability, robustness, and efficiency are often as important as peak predictive accuracy \cite{rodemann_explaining_2026,islam_explainable_2025,he_survey_2025}. Recent work has therefore explored explainable HPO by adapting feature-attribution techniques, most notably Shapley values, to analyze hyperparameter influence \cite{wever_hypershap_2025,boukrouh_comparative_2024,wang_solutions_2022}. Although informative, these approaches are typically post-hoc, model-specific, and computationally demanding \cite{verhaeghe_powershap_2023,mu_shrinkhpo_2024}.

Shapley values, originating in cooperative game theory \cite{kuhn_17_1953}, provide a principled mechanism for allocating global performance contributions while accounting for interactions. Their impact spans explainable AI, feature attribution, and data valuation \cite{muschalik_shapiq_2024,liu_shapley_2023,wang_rethinking_2024}. In global sensitivity analysis (GSA), Shapley Effects overcome key limitations of variance-based Sobol’ indices, particularly under nonlinearities, interactions, and correlated inputs \cite{owen_shapley_2017,benoumechiara_shapley_2019,iooss_shapley_2019,plischke_computing_2021,stein_shapley_2023}. These properties are especially relevant for hyperparameter spaces in deep learning, which are high-dimensional and interaction-dominated \cite{pheulpin_uncertainty_2022,iooss_review_2015}.

In this manuscript, we propose a game-theoretic framework for hyperparameter--objective interaction analysis that leverages \emph{Shapley spectra} and \emph{Pareto front} analysis to characterize player influence in a cooperative game formulation applicable to machine learning models across different tasks. These spectra expose architecture-dependent sensitivity patterns, identifying hyperparameters that dominate performance and those with negligible impact that can be fixed or deprioritized~\cite{owen_shapley_2017,benoumechiara_shapley_2019}. By leveraging input--output mapping-based game-theoretic global sensitivity analysis (GSA) techniques, Shapley spectra can be estimated efficiently without requiring exhaustive retraining~\cite{blatman_efficient_2010,duan_derivative-based_2023}.

The proposed framework provides practitioners with a general analysis approach that yields reusable and transferable prior knowledge across similar models and tasks while requiring minimal additional computational overhead. The remainder of this paper is organized as follows: Section~\ref{sec:method} introduces the proposed methodological framework, Section~\ref{sec:experiment} evaluates the approach through a set of experimental studies, Section~\ref{sec:analysis} discusses the key takeaways and limitations of the proposed framework, and Section~\ref{sec:conclusion} concludes the paper. The codebase and datasets used in this study are publicly available at \texttt{https://github.com/NyiNyi-14/GT\_MOHA.git}.

%
%
%
\section{METHODOLOGICAL FRAMEWORK} 
\label{sec:method}
\begin{figure}[ht]
    \centering
    \includegraphics[clip, trim= 0 665 305 0,width=\linewidth]{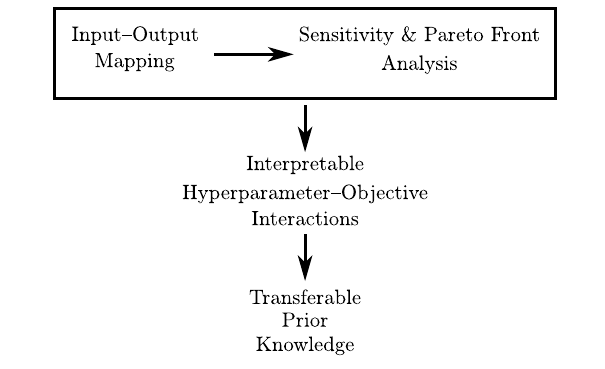}
    \caption{Overview of the proposed framework for extracting interpretable hyperparameter--objective interactions and transferable prior knowledge from input--output mappings.}
    \label{fig:EMOHO_diag}
\end{figure}
In the proposed framework, each tunable component of the learning algorithm is modeled as a player in a cooperative game, where the players correspond to hyperparameters in the context of neural network optimization. The overall workflow of the proposed framework is illustrated in Fig.~\ref{fig:EMOHO_diag}. To construct the input--output mappings between players and objectives, a coarse-grained search is employed in this work. Nevertheless, the proposed framework is not restricted to coarse-grained search and can also be applied to alternative input--output mapping scenarios. The resulting mappings are stored in lookup tables, after which Shapley Effects and non-dominated sets are employed for sensitivity and Pareto front analysis. Finally, interpretable interactions between the players and the objectives are extracted as transferable prior knowledge.


%
%
%
\subsection{Game-Theoretic Formulation}
This step defines the game-theoretic formulation of the proposed framework, including the number of players and the values assigned to each player. Regardless of the total number of players, we recommend using a small set of representative values, typically lower, nominal, and upper bounds, to balance computational efficiency and sensitivity characterization. At this stage, fine-grained hyperparameter search is unnecessary, as the objective is to capture global trends rather than precise optima. Let $\mathcal{P} = \{p_1, p_2, \ldots, p_n\}$ denote the set of players,
\begin{subequations}
\begin{align}
p_k &\in \mathcal{A}_k = \{q_{k,1}, q_{k,2}, \ldots, q_{k,m_k}\}, \\[2pt]
\Phi &= \mathcal{A}_1 \times \mathcal{A}_2 \times \cdots \times \mathcal{A}_n \subseteq \mathbb{R}^n, \\[2pt]
\lvert \Phi \rvert &= \prod_{k=1}^{n} m_k, \\[2pt]
\bm{\phi}_i &= (p_1^{(i)}, p_2^{(i)}, \ldots, p_n^{(i)}) \in \Phi, \\[2pt]
\mathcal{Y}_i &= (y_{i,1}, y_{i,2}, \ldots, y_{i,J}) = f(\bm{\phi}_i), \\[2pt]
\mathcal{D} 
&= \{ (\bm{\phi}_i, \mathcal{Y}_i) \mid \bm{\phi}_i \in \Phi \}_{i = 1}^{\lvert \Phi \rvert},
\end{align}
\label{eq:proxy}
\end{subequations}
\vspace{-5mm}

\noindent
where $\mathcal{A}_k$ denotes the discrete value set of the $k$-th player, $q_{k,m_k}$ represents the $m_k$-th candidate value (action) assigned to player $p_k$, $\Phi$ is the Cartesian product of all such sets defining the player configuration space, $\phi_i \in \Phi$ represents an individual configuration, and $(y_{i,1}, \ldots, y_{i,J})$ denotes the vector of application-dependent objectives produced by evaluating $\phi_i$. The lookup table $\mathcal{D}$, which stores all evaluated player configurations together with their corresponding objective outputs, is constructed at the end of the coarse-grained search.
%
%
%
%
%
%
\subsection{Monte Carlo-Based Sobol--Shapley Sensitivity Analysis}
\paragraph*{Coalition Construction}
First, all possible coalitions are enumerated as defined in \eqref{eq:holdings}, and for each coalition, the corresponding conditional subsets are constructed by fixing the selected players.
\begin{subequations}
\begin{align}
\mathcal{C}_k 
&= \{ \mathcal{S} \subseteq \mathcal{P} \mid |\mathcal{S}| = k \}, 
\qquad k = 1,\ldots,n, \\[2pt]
|\mathcal{C}_k| &= \binom{n}{k},
\qquad
|\mathcal{C}| = \sum_{k=1}^{n} \binom{n}{k} = 2^n - 1, \\[2pt]
\mathcal{G}_\mathcal{S} &= \prod_{i \in \mathcal{S}} \mathcal{G}_i  = \{ \bm{g}_\mathcal{S}^{(1)}, \ldots, \bm{g}_\mathcal{S}^{(|\mathcal{G}_\mathcal{S}|)} \}, \\[2pt]
\mathcal{I}(\mathcal{S},\bm{g}_\mathcal{S}) 
&= \{ r \in \{1,\ldots,{\lvert \Phi \rvert}\} \mid \phi_{r,\mathcal{S}} = \bm{g}_\mathcal{S} \}, \\[2pt]
\mathcal{D}(\mathcal{S},\bm{g}_\mathcal{S})
&= \{ (\bm{\phi}_r, \mathcal{Y}_r) \mid r \in \mathcal{I}(\mathcal{S},\bm{g}_\mathcal{S}) \}.
\end{align}
\label{eq:holdings}
\end{subequations}
\vspace{-5mm}

\noindent
Here, $\mathcal{C}_k$ denotes the set of all coalitions of size $k$, where each coalition $\mathcal{S}$ specifies a subset of players to be held fixed. $\mathcal{G}_{\mathcal{S}}$ is the Cartesian product of value sets for all players in $\mathcal{S}$ and $\mathcal{G}_i$ denotes the discrete value set assigned to player $i$. For a given assignment $\bm{g}_{\mathcal{S}} \in \mathcal{G}_{\mathcal{S}}$, 
$\mathcal{I}(\mathcal{S}, \bm{g}_{\mathcal{S}})$ denotes the index set of lookup-table entries whose configurations match $\bm{g}_{\mathcal{S}}$ on $\mathcal{S}$, and 
$\mathcal{D}(\mathcal{S}, \bm{g}_{\mathcal{S}})$ is the corresponding conditional subset of the lookup table. 
%
%
%
\paragraph*{Monte Carlo Estimation}
To avoid exhaustive enumeration of the full search space, a coarse-grained search is adopted in place of a fine-grained exploration by bounding player values, yielding a lookup table with on the order of \(\sim10^2\) configurations. Consequently, Monte Carlo (MC) estimation is employed to approximate Sobol indices using a larger number of samples, rather than relying on exact computations from the limited discrete lookup table. Then, conditional variances are estimated by sampling over $\mathcal{D}(\mathcal{S},\bm{g}_\mathcal{S})$, as detailed below.
\begin{subequations}
    \begin{align}
        r^t &\stackrel{\text{i.i.d.}}{\sim} \mathcal{U}(\mathcal{I}(\mathcal{S},\bm{g}_\mathcal{S})),
        \quad t=1,\ldots,N, \\[2pt]
        \bm{y}_{\mathrm{MC}} &= [f(\bm{\phi}_{r^1}), f(\bm{\phi}_{r^2}), \ldots, f(\bm{\phi}_{r^N})], \\[2pt]
       \widehat{\mu}(\mathcal{S},\bm{g}_{\mathcal{S}}) &= \widehat{\mathbb{E}}\!\left[Y \mid \bm{G}_{\mathcal{S}}=\bm{g}_{\mathcal{S}}\right] = \frac{1}{N}\sum_{t=1}^{N} \bm{y}_{\mathrm{MC}}^{(t)},
    \end{align}
    \label{eq:MC_estimator}
\end{subequations} 
\vspace{-3mm}

\noindent
where \(r^{(t)}\) denotes MC indices drawn uniformly from $\mathcal{I}(\mathcal{S},\bm{g}_{\mathcal{S}})$ for $N$ samples, with $N = 10^4$ used in this work, $\bm{y}_{\mathrm{MC}}$ are the corresponding model outputs and $\widehat{\mu}(\mathcal{S},\bm{g}_{\mathcal{S}})$ denotes the MC estimator of the conditional mean.
%
%
%
\paragraph*{Sobol Variance Decomposition}
Since player sensitivity is normally characterized by strong interactions and nonlinear effects, global sensitivity analysis is particularly appropriate when gradient information is often unavailable. Among available approaches, Sobol indices are of interest because they quantify, in a variance-based and model-agnostic manner, the proportion of output uncertainty explained by fixing a subset of players. In this work, first-order Sobol indices derived from the Sobol variance decomposition in \eqref{eq:sobol_ind} are employed \cite{iooss_shapley_2019}.
\begin{equation}
    v(\mathcal{S}) =
\frac{\widehat{\mathrm{Var}}\big(\mathbb{E}[\mathcal{Y} \mid \bm{G}_{\mathcal{S}}]\big)}{\mathrm{Var}(\mathcal{Y})},
\label{eq:sobol_ind}
\end{equation}
where $v(\mathcal{S})$ denotes the coalition value following the Sobol coalition-worth formulation,
with $v(\emptyset)=0$ corresponding to the empty set. $\mathrm{Var}(\mathcal{Y})$ is the global output variance, and $\widehat{\mathrm{Var}}\big(\mathbb{E}[\mathcal{Y} \mid \bm{G}_{\mathcal{S}}]\big)$ is the variance of the conditional expectation which are estimated using the MC procedure described in \eqref{eq:global_var}.
\begin{subequations}
    \begin{align}
        \mathrm{Var}(\mathcal{Y}) =
\frac{1}{|\Phi|}\sum_{r=1}^{|\Phi|}\left(\bm{y}_r-\bar{y}\right)^2, \quad \bar{y}=\frac{1}{|\Phi|}\sum_{r=1}^{|\Phi|}\bm{y}_r, \\[2pt]
\widehat{\mu}_{\mathcal{S}} = \frac{1}{|\mathcal{G}_{\mathcal{S}}|}\sum_{\bm{g}_{\mathcal{S}}\in\mathcal{G}_{\mathcal{S}}} \widehat{\mu}(\mathcal{S},\bm{g}_{\mathcal{S}}), \\[2pt]
\widehat{\mathrm{Var}}\big(\mathbb{E}[\mathcal{Y} \mid \bm{G}_{\mathcal{S}}]\big) =
\frac{1}{|\mathcal{G}_{\mathcal{S}}|}
\sum_{\bm{g}_{\mathcal{S}}\in\mathcal{G}_{\mathcal{S}}}
\left(\widehat{\mu}(\mathcal{S},\bm{g}_{\mathcal{S}})-\widehat{\mu}_{\mathcal{S}}\right)^2.
    \end{align}
    \label{eq:global_var}
\end{subequations}
%
%
%
\paragraph*{Shapley Effect Estimation}
While Sobol indices quantify the contribution of subsets of players to the total output uncertainty, they do not yield a unique attribution of importance to individual players in the presence of interactions. To address this limitation, Shapley Effects are employed, as defined in~\eqref{eq:shapley}, to fairly redistribute the coalition variance across individual players by averaging marginal contributions over all possible coalition orderings \cite{kuhn_17_1953}:
\begin{equation}
\mathbb{S}_i^{v_j}
=
\sum_{\mathcal{S}\subseteq \mathcal{P}\setminus\{i\}}
\frac{|\mathcal{S}|!\,(n-|\mathcal{S}|-1)!}{n!}
\left[
v_j(\mathcal{S}\cup\{i\}) - v_j(\mathcal{S})
\right],
\label{eq:shapley}
\end{equation}
where $\mathbb{S}_i^{v_j}$ denotes the Shapley value of player $i$, with respect to objective $j$, $v_j : 2^{\mathcal{P}} \rightarrow \mathbb{R}$ is the objective-specific coalition value function and $n = |\mathcal{P}|$ is the total number of players. Here, $v(\mathcal{S}\cup\{i\})$ denotes the value of the coalition obtained by adding player $i$ to coalition $\mathcal{S}$, while $v(\mathcal{S})$ denotes the value of the coalition excluding player $i$. This Sobol--Shapley combination yields interaction aware and mode agnostic global sensitivity measures suitable for player analysis.
\begin{equation}
    \sum_{i\in\mathcal{P}} \mathbb{S}_i^{v_j} = 1.
    \label{eq:shapley_sum}
\end{equation}
Since the Shapley Effects $\mathbb{S}_i^{v_j}$ analytically sum to unity for each objective, as shown in~\eqref{eq:shapley_sum}, they provide a normalized measure of player importance across objectives. Consequently, the relative influence of each player on the objectives can be quantified and compared directly. Furthermore, as defined in~\eqref{eq:fix_vary}, a player \(i\) is categorized as less sensitive if its Shapley Effect remains below a user-defined threshold \(\tau\) for all objectives. Otherwise, the player is considered highly sensitive with respect to the game objectives.
\begin{equation}
\label{eq:fix_vary}
p_i =
\begin{cases}
\text{Non-influential}, & \max\limits_{j \in \{1,\dots,J\}} \mathbb{S}_i^{v_j} < \tau,\\
\text{Influential}, & \text{otherwise}.
\end{cases}
\end{equation}
This classification facilitates identifying which players are most influential with respect to each objective and which players exhibit negligible impact across the objectives.
%
%
%
\subsection{Pareto-Optimal Player Set Identification}

During the coarse-grained search, each configuration is evaluated only once, and the resulting performance can be noisy due to the inherent stochasticity of neural network training. To mitigate this effect and avoid selecting configurations that benefit from occasional ``lucky'' runs, Pareto optimality~\cite{deb2011multi} is employed to identify non-dominated player sets from the stored lookup table $\mathcal{D}$. The resulting Pareto front provides insight into the trade-offs associated with different player configurations before selecting a final candidate set. Furthermore, this analysis enables early-stage model evaluation by revealing whether the model exhibits sufficient performance potential under different hyperparameter combinations prior to fine-grained hyperparameter tuning.
\begin{definition}[Pareto dominance]
Let \(\{\mathcal{Y}^{(i)} \}_{i=1}^{N}\) be the set of evaluated solutions of $\mathcal{D}$ from \eqref{eq:proxy}, where \(\mathcal{Y}^{(i)} \in \mathbb{R}^{J}\) is the objective vector for solution \(i\) across \(J\) objectives. Each solution corresponds to the objective vector induced by a distinct player set $\bm{\phi}_i$ evaluated during proxy training. Let the \(j\)-th objective be denoted by \(y^{(i)}_j\), with direction
\(\delta_j \in \{\text{min},\text{max}\}\). Since objectives can have different optimization directions, a simple sign transformation is applied to convert all objectives into a minimization form. Specifically, for the \(j\)-th objective, we define:
\begin{equation}
s_j =
\begin{cases}
\;\;1, & \delta_j = \text{min},\\
-1, & \delta_j = \text{max},
\end{cases}
\qquad j=1,\ldots,J,
\end{equation}
and the transformed objective vector:
\begin{equation}
\tilde{\mathcal{Y}}^{(i)} = \big(s_1 y^{(i)}_1,\ldots,s_J y^{(i)}_J\big) \in \mathbb{R}^{J},
\quad i=1,\ldots,N,
\end{equation}
so that all objectives are cast into a minimization form. The solution $k$ is said to dominate the solution $l$ if
\begin{subequations}
\begin{align}
\tilde{\mathcal{Y}}^{(k)} &\preceq \tilde{\mathcal{Y}}^{(l)}
\;\;\text{and}\;\;
\tilde{\mathcal{Y}}^{(k)} \neq \tilde{\mathcal{Y}}^{(l)}, 
\label{eq:pareto_dom_vector} \\
\tilde{y}^{(k)}_j &\le \tilde{y}^{(l)}_j \;\; \forall j\in\{1,\ldots,J\}, \quad
\exists \; j \;\text{s.t.}\; \tilde{y}^{(k)}_j < \tilde{y}^{(l)}_j.
\label{eq:pareto_dom_component}
\end{align}
\end{subequations}
Finally, a solution \(k\) is defined as \textbf{Pareto-optimal (non-dominated)} set if there does not exist any \(l\neq k\) that dominates it:
\begin{equation}
k \in \mathcal{P}^\star
\iff
\nexists\, l\in\{1,\ldots,N\}\setminus\{k\}
\;\; \text{s.t.} \;\;
\tilde{\mathcal{Y}}^{(l)} \prec \tilde{\mathcal{Y}}^{(k)}.
\label{eq:pareto_set}
\end{equation}
\end{definition}
The set \(\mathcal{P}^\star\) constitutes the Pareto (non-dominated) front in the transformed objective space and provides information for early-stage model evaluation by characterizing the attainable trade-offs among the objectives.
%
%
%
\section{CASE STUDIES \& EXPERIMENTAL RESULTS}
\label{sec:experiment}
To evaluate the proposed framework, we conduct experiments on three distinct machine learning applications spanning different model classes and task types. 
%
%
%
\subsection{Experimental Setup}
We consider (i) a physics-informed neural network (PINN), (ii) a convolutional neural network (CNN), and (iii) a deep neural network (DNN), each applied to a different task. 
Across all applications, each hyperparameter is treated as a player in the cooperative game formulation. For each model, the player set $\mathcal{P} = \{p_1, p_2, \ldots\}$ follows the ordering of hyperparameters listed in the corresponding coarse-grained search tables. In the absence of prior knowledge, these hyperparameter ranges are chosen based on common practice.
%
%
%
\subsubsection{Physics-Informed Neural Network: Dual Spring--Mass--Damper System}
\begin{figure}
\centering
\includegraphics[clip, trim=30 720 260 30, width=0.9\linewidth]{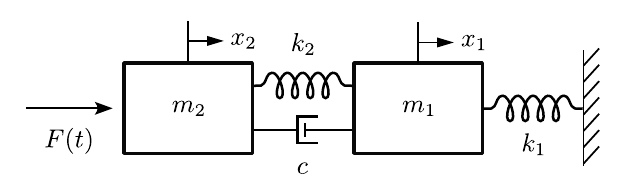}
\caption{Dual spring--mass--damper system.}
\label{fig:spring-mass-diag}
\vspace{-2mm}
\end{figure}

The dual spring--mass--damper system is used as a case study for the PINN~\cite{RAISSI2019686} experiments, where the $R^2$ metric is employed to evaluate the accuracy of the predicted mass displacements against the ground truth trajectories. The corresponding hyperparameter ranges are summarized in Table~\ref{tab:pinn_hyperparams}. The number of training steps is fixed at $15 \times 10^{3}$, and a linear activation function is employed in the output layer. Since three objectives are considered, namely the prediction accuracies of masses~1 and~2 together with the training duration, the problem naturally constitutes a multi-objective optimization setting. Additional details regarding the system dynamics, including the schematic diagram and physical parameters, are provided in Fig.~\ref{fig:spring-mass-diag} and Table~\ref{tab:SPM_para}.
\begin{table}
\caption{Coarse-grained hyperparameter search for PINN.}
  \label{tab:pinn_hyperparams}
  \begin{center}
    \begin{small}
      \begin{sc}
        \begin{tabular}{l l}
          \toprule
          Hyperparameter & Candidate Values \\
          \midrule
          Learning rate              & $\{10^{-3},\,10^{-4}\}$ \\[1pt]
          Number of layers               & $\{2,\,4\}$ \\[1pt]
          Neurons per layer           & $\{64,\,128\}$ \\[1pt]
          Physics loss weight ($\lambda_{\mathrm{phys}}$) & $\{10^{-4},\,10^{-2}\}$ \\[1pt]
          Data loss weight ($\lambda_{\mathrm{data}}$)    & $\{1,\,10\}$ \\[1pt]
          Physics collocation points  & $\{60,\,120,\,240\}$ \\[1pt]
          Activation function                   & \{Tanh, ReLU, Sine\} \\
          \bottomrule
        \end{tabular}
      \end{sc}
    \end{small}
  \end{center}
  \vskip -0.1in
\end{table}
\begin{table}
\caption{Parameters of the dual spring--mass--damper system.}
\label{tab:SPM_para}
\centering
\begin{small}
\begin{tabular}{lll}
\toprule
Description & Value & Unit \\
\midrule
Mass of first body ($m_1$)        & 1   & kg \\
Mass of second body ($m_2$)       & 2   & kg \\
Spring stiffness ($k_1$)          & 5   & N/m \\
Spring stiffness ($k_2$)          & 10  & N/m \\
Damping coefficient ($c$)         & 0.5 & N\,s/m \\
External force ($F(t)$)           & $\sin(t)$ & N \\
\bottomrule
\end{tabular}
\end{small}
\end{table}
The governing equations of the system are given by:
\begin{subequations}
    \begin{align}
    m_1 \ddot{x}_1 - c(\dot{x}_2 - \dot{x}_1) - k_2 x_2 + (k_1 + k_2)x_1 &= 0, \\[2pt]
    m_2 \ddot{x}_2 + c(\dot{x}_2 - \dot{x}_1) + k_2(x_2 - x_1) &= F(t).
    \end{align}
\end{subequations}
%
%
%
\subsubsection{Convolutional Neural Network: CIFAR-100 Classification}
The CIFAR-100 dataset is used to evaluate the CNN on a multi-class image classification task. With 100 classes, each containing 600 color images of size $32 \times 32$, CIFAR-100 represents a relatively large-scale benchmark and is well suited for evaluating the proposed framework. Additional details regarding the dataset are provided in~\cite{Krizhevsky2009LearningML}.
\begin{table}[ht]
  \caption{Coarse-grained hyperparameter search for CNN.}
  \label{tab:cnn_hyperparams}
  \begin{center}
    \begin{small}
      \begin{sc}
        \begin{tabular}{l l}
          \toprule
          Hyperparameter & Candidate Values \\
          \midrule
          Learning rate                & $\{10^{-3},\,10^{-4}\}$ \\[1pt]
          Batch size                   & $\{16,\,64\}$ \\[1pt]
          Convolutional filters 1 & $\{32,\,64\}$ \\[1pt]
          Convolutional filters 2 & $\{64,\,128\}$ \\[1pt]
          Convolutional filters 3 & $\{128,\,256\}$ \\[1pt]
          Kernel size                  & $\{3,\,5\}$ \\[1pt]
            Activation function          & \{ReLU, Tanh, SiLU\} \\[1pt]
        Batch Normalization                   & \{True,\,False\} \\

          \bottomrule
        \end{tabular}
      \end{sc}
    \end{small}
  \end{center}
  \vskip -0.1in
\end{table}
The CNN architecture is configured as summarized in Table~\ref{tab:cnn_hyperparams} with the number of training epochs fixed to 20 for all runs. The activation function is varied only within the convolutional layers, while the activation applied after flattening is fixed to ReLU, and the output layer employs a linear activation across all configurations. For the CNN classification task, a multiclass linear classifier is used at the output layer, and class predictions are obtained by applying an argmax operation to the output logits, without an explicit sigmoid or softmax activation during inference. Prediction accuracy and training duration are treated as the objective outputs in the multi-objective optimization setting.
%
%
%
\subsubsection{Deep Neural Network: Adult Income Classification}
The Adult Income dataset is used for the DNN binary classification task, where the network predicts whether an individual’s annual income exceeds \$50K based on 14 input features. Dataset details are provided in~\cite{becker_adult_1996}. 
\begin{table}[ht]
  \caption{Coarse-grained hyperparameter search for DNN.}
  \label{tab:dnn_hyperparams}
  \begin{center}
    \begin{small}
      \begin{sc}
        \begin{tabular}{l l}
          \toprule
          Hyperparameter & Candidate Values \\
          \midrule
          Learning rate            & $\{10^{-3},\,10^{-4}\}$ \\[1pt]
          Batch size               & $\{32,\,64,\,128\}$ \\[1pt]
          Number of layers         & $\{2,\,4\}$ \\[1pt]
          Neurons per layer        & $\{64,\,128\}$ \\[1pt]
          Activation function      & $\{\text{ReLU},\,\text{Tanh},\,\text{SiLU}\}$ \\[1pt]
        Batch normalization      & $\{\text{True},\,\text{False}\}$ \\[1pt]
          Dropout rate             & $\{0.0,\,0.3\}$ \\[1pt]

          \bottomrule
        \end{tabular}
      \end{sc}
    \end{small}
  \end{center}
  \vskip -0.1in
\end{table}
The coarse-grained search space is summarized in Table~\ref{tab:dnn_hyperparams}, with the number of training epochs fixed to 50 for all runs. A linear activation is used at the output layer across all configurations, and a binary logistic classifier is employed, trained using a sigmoid-based binary cross-entropy loss. Prediction accuracy and training duration are considered as the two competing objectives in this multi-objective optimization case study.
%
%
%
\subsection{Hyperparameter--Objective Interaction Analysis}
The game-theoretic formulation for each application is summarized in Table~\ref{tab:emoho_games}. For each network, a coarse-grained search is conducted over the corresponding player set, and each evaluated configuration, together with its associated objective values, is stored in a lookup table. Sample lookup tables are provided in Tables~\ref{tab:pinn_lookup_sample}--\ref{tab:dnn_lookup_sample} in Appendix~\ref{Asec:lookup}.
\begin{table*}[ht]
\caption{Game-theoretic formulation of the search across learning tasks.}
  \label{tab:emoho_games}
  \begin{center}
    \begin{small}
      \begin{sc}
        \begin{tabular}{l c c c c}
          \toprule
          Application & Set of Players, $\mathcal{P}$ & Objectives, $\mathcal{Y}$ &  Configurations, $\lvert \Phi \rvert$ & Remarks \\
          \midrule
          Pinn & $\{p_1,\,p_2,\,\ldots,\,p_7\}$ & $\{y_1,\,y_2,\,t\}$ & 288 & $7$-player, 3-objective game \\[2pt]
          Cnn  & $\{p_1,\,p_2,\,\ldots,\,p_8\}$ & $\{y_1,\,t\}$    &   384  & $8$-player, 2-objective game \\[2pt]
          Dnn  & $\{p_1,\,p_2,\,\ldots,\,p_7\}$ & $\{y_1,\,t\}$    & 288  & $7$-player, 2-objective game \\
          \bottomrule
        \end{tabular}
      \end{sc}
    \end{small}
  \end{center}
  \vskip -0.1in
\end{table*}

\begin{figure*}[ht]
    \centering
    \includegraphics[clip, trim= 5 640 5 40,width=\linewidth]{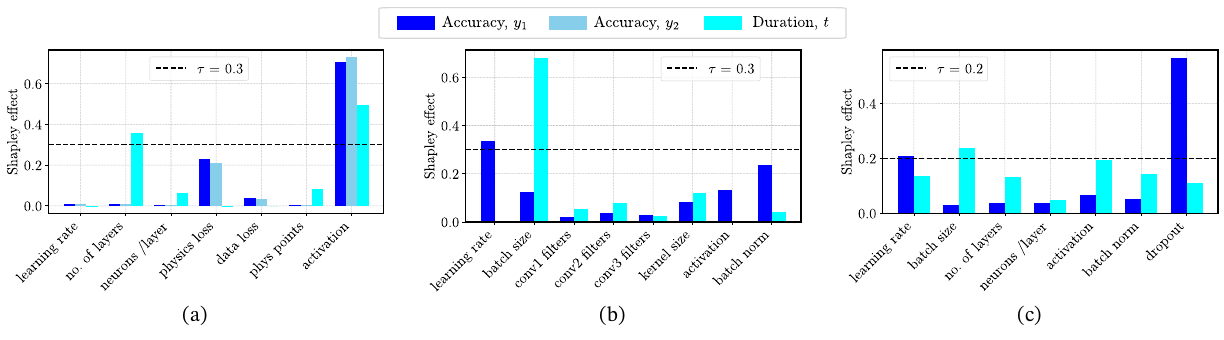}
    \caption{Shapley-based global sensitivity analysis for (a) PINN, (b) CNN, and (c) DNN, with the corresponding threshold~$\tau$.}
    \label{fig:sensitivity}
\end{figure*}

\paragraph{Global sensitivity analysis} The Shapley Effect is estimated according to \eqref{eq:shapley}, and the resulting global sensitivity of each player in the game with respect to the corresponding objectives is visualized in Fig.~\ref{fig:sensitivity}. The selection of the threshold value \(\tau\) from \eqref{eq:fix_vary} is user-defined and application-dependent. In general, as the number of players in the game increases, individual sensitivity values tend to be more dispersed. The specific value of \(\tau\) adopted in this work is chosen for experimental purposes to demonstrate the proposed framework, and no claim of optimality is made regarding this selection and alternative threshold values may be selected to reflect different application requirements or user preferences.
\begin{equation}
\label{eq:shapley_sum2}
     \sum_{i\in\mathcal{P}} \mathbb{S}_i^{y_1} \approx 
 \sum_{i\in\mathcal{P}} \mathbb{S}_i^{y_2} \approx
 \sum_{i\in\mathcal{P}} \mathbb{S}_i^{t} \approx
 1.
\end{equation}
The Shapley effects are normalized and sum to unity for each objective, as shown in \eqref{eq:shapley_sum2}. As illustrated in Fig.~\ref{fig:sensitivity}(a) and (b), several hyperparameters exhibit negligible sensitivity across all objectives and therefore receive less emphasis during subsequent optimization or fine-tuning stages. Consequently, these hyperparameters can be assigned reduced search spaces by either fixing them to a single value or restricting them to a small set of candidate values. In contrast, Fig.~\ref{fig:sensitivity}(c) shows that the sensitivities of most hyperparameters are relatively evenly distributed, suggesting that the DNN binary classification task considered in this study is comparatively less sensitive to hyperparameter variations. Therefore, extensive hyperparameter fine-tuning is expected to provide only limited improvement in the objectives for the considered model and dataset.

\paragraph{Pareto front set identification}
\begin{figure*}[ht]
    \centering
    \includegraphics[clip, trim= 5 650 5 30 30,width=\linewidth]{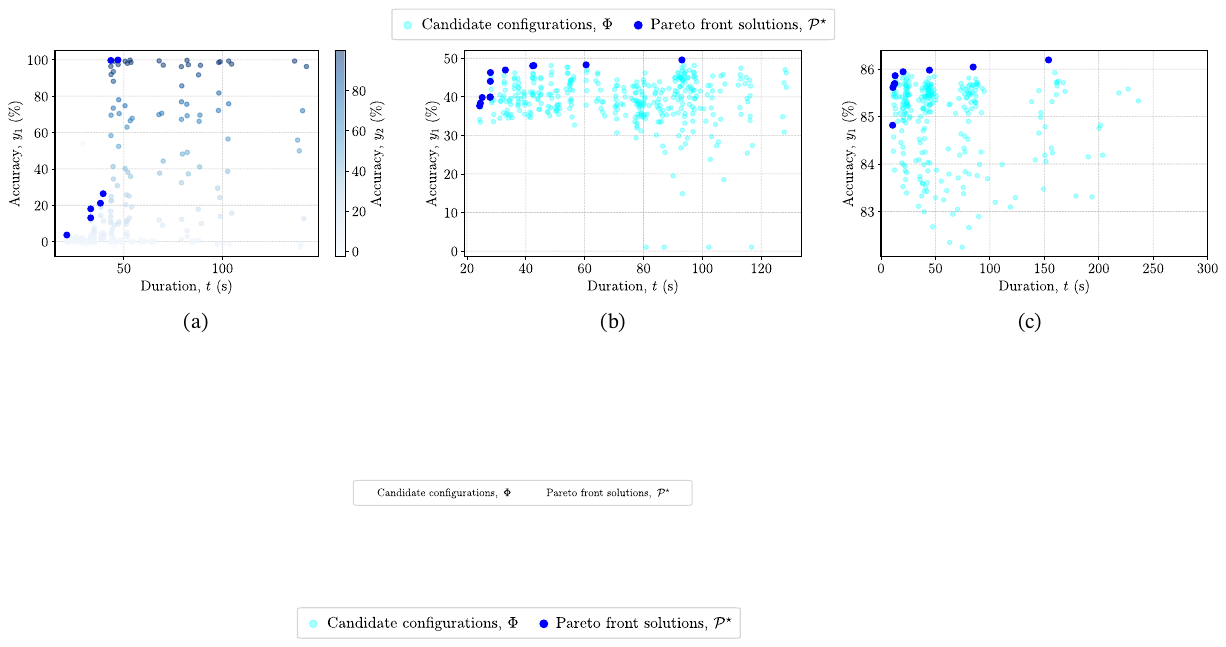}
    \caption{Identification of Pareto front solutions among candidate configurations for (a) PINN, (b) CNN, and (c) DNN.}
    \label{fig:pareto_iden}
\end{figure*}
The Pareto-optimal set is identified according to \eqref{eq:pareto_set}, and the results are visualized in Fig.~\ref{fig:pareto_iden} which provides an early-stage evaluation of model performance for the corresponding tasks. In Fig.~\ref{fig:pareto_iden}(a), the PINN accuracies $y_1$ and $y_2$ span a wide range, from near zero up to approximately $99\%$, while the training duration also varies significantly. This indicates substantial optimization potential, but also highlights a pronounced trade-off between accuracy and computational cost. In contrast, Fig.~\ref{fig:pareto_iden}(b) shows that the CNN used for multi-class classification attains a maximum accuracy of approximately $50\%$, largely independent of training duration suggesting that the model has reached a saturated performance regime. Similarly, Fig.~\ref{fig:pareto_iden}(c) demonstrates that the accuracy of the DNN binary classification task is confined to a relatively narrow range (approximately $83\%$ to $86\%$), indicating weak sensitivity to hyperparameter variations for the selected dataset. 

Therefore, the combined global sensitivity analysis and Pareto front evaluation suggest that the PINN is well suited for predicting the dynamics of the dual spring--mass--damper system. In this case, local refinement can be performed by selecting balanced trade-off candidate configurations from the Pareto front and conducting further search in the surrounding region. In contrast, the CNN with a linear classifier applied to the CIFAR-100 multi-class classification task is not expected to achieve substantial improvement in the objectives through additional hyperparameter fine-tuning, since the Pareto front remains concentrated in a relatively low-accuracy region across all candidate configurations. Similarly, the DNN binary classification task on the Adult Income dataset exhibits comparatively low sensitivity to hyperparameter variations. These observations suggest that the proposed framework can provide early-stage model evaluation by revealing the foreseeable performance potential of a model for a given task prior to extensive hyperparameter optimization.
%
%
%
\section{KEY TAKEAWAYS \& LIMITATIONS OF THE PROPOSED FRAMEWORK}
\label{sec:analysis}
This work presents a sensitivity-based framework for analyzing hyperparameter-–objective interactions in machine learning models rather than a new hyperparameter optimization (HPO) algorithm. Consequently, comparisons with existing HPO baselines are outside the scope of this work. As illustrated in Fig.~\ref{fig:sensitivity}, the influence of a hyperparameter is highly application-dependent, where even the same hyperparameter may exhibit substantially different sensitivities across models and objectives. For example, the learning rate may strongly influence the prediction accuracy of one model while having comparatively lower impact in another. Similar observations are found for the remaining hyperparameters, highlighting that hyperparameter importance cannot be generalized across different learning tasks. The resulting interaction analysis provides interpretable insights into the relative importance of design variables and may also serve as prior knowledge for subsequent hyperparameter optimization. 

The proposed framework currently relies on a coarse-grained grid search to construct the input--output mappings required for the Shapley-based sensitivity analysis. Consequently, the resulting interaction analysis depends on the selected search space and the discretization of the hyperparameter values. In this work, the coarse-grained search serves primarily as a proof of concept for demonstrating the proposed analysis framework rather than as an optimal hyperparameter search strategy. Future research will investigate principled approaches for selecting representative sampling points and extend the framework to operate with more general input--output mappings obtained from alternative optimization strategies, such as random search, Bayesian optimization, or surrogate-based optimization methods.
%
%
%
\section{CONCLUSION}
\label{sec:conclusion}
This manuscript presents a framework for hyperparameter--objective interaction analysis that enables the interpretation of which hyperparameters are most influential with respect to different objectives and which hyperparameters require less attention. The proposed framework employs sensitivity analysis and Pareto front evaluation on the input--output mappings between players and objectives in a cooperative game formulation, thereby providing transferable prior knowledge regarding the considered model and task. As future work, the proposed framework will be extended to alternative search and sampling algorithms and applied to a broader range of systems.
%
%
%

\bibliographystyle{IEEEtran}
\bibliography{references}  
%
%
%

\clearpage
\onecolumn
\appendix

\renewcommand{\thefigure}{A\arabic{figure}}
\setcounter{figure}{0}
\renewcommand{\thetable}{A\arabic{table}}
\setcounter{table}{0}

\section{Appendix}
\label{sec:appendix}

\subsection{Coarse-Grained Search Results for Each Model}
\label{Asec:lookup}
The evaluated objectives ($y_1, y_2, \ldots$) and their corresponding players ($p_1, p_2, \ldots$) from grid search are stored in a lookup table, and sample lookup tables for all models are summarized in Tables~\ref{tab:pinn_lookup_sample}--\ref{tab:dnn_lookup_sample}.
\begin{table*}[ht]
\caption{Sample entries from the PINN coarse-grained search lookup table.}
\label{tab:pinn_lookup_sample}
\centering
\begin{small}
\begin{tabular}{c c c c c c c c c c c}
\toprule
\multirow{2}{*}{Run} &

$p_1$ &
$p_2$ &
$p_3$ &
$p_4$ &
$p_5$ &
$p_6$ &
$p_7$ &
$y_1$ &
$y_2$ &
$t$ \\[2pt]

 & $(\mathrm{lr})$ &
$(n_{\mathrm{layers}})$ &
$(n_\mathrm{neurons})$ &
$(\lambda_{\mathrm{phys}})$ &
$(\lambda_{\mathrm{data}})$ &
$(n_{\mathrm{phys}})$ &
(Activation) &
$(R^2_{x_1})$ &
$(R^2_{x_2}$) &
(sec) \\
\midrule
1 & 0.001 & 2 & 64 & $10^{-4}$ & 1 & 60  & Tanh & $-0.001$ & $0.0007$ & 28.61 \\[2pt]

2 & 0.001 & 2 & 64 & $10^{-4}$ & 1 & 60  & ReLU & $0.0006$  & $0.007$ & 21.31 \\[2pt]

3 & 0.001 & 2 & 64 & $10^{-4}$ & 1 & 60  & Sine & 0.2253 & 0.2608 & 43.52 \\[2pt]

\vdots & \vdots & \vdots & \vdots & \vdots & \vdots & \vdots & \vdots & \vdots & \vdots & \vdots \\[2pt]

288 & 0.0001 & 4 & 128 & $10^{-2}$ & 10 & 240 & Sine & 0.7202 & 0.7294 & 140.73 \\
\midrule
&   &   &  &  &  &  &   &\multicolumn{2}{c}{\sc{total duration}} & $\approx$ 4-hr \\
\bottomrule
\end{tabular}
\end{small}
\vspace{-0.08in}
\end{table*}
\begin{table*}[ht]
\caption{Sample entries from the CNN coarse-grained search lookup table.}
\label{tab:cnn_lookup_sample}
\centering
\begin{small}
\begin{tabular}{c c c c c c c c c c c}
\toprule
\multirow{2}{*}{Run} &

$p_1$ &
$p_2$ &
$p_3$ &
$p_4$ &
$p_5$ &
$p_6$ &
$p_7$ &
$p_8$ &
$y_1$ &
$t$ \\[2pt]

 & 
($\mathrm{lr}$) &
(Batch) &
(Conv$_1$) &
(Conv$_2$) &
(Conv$_3$) &
(Kernel) &
(Activation) &
(Batch Norm) &
(Accuracy, \% )&
(sec) \\
\midrule
1 & 0.001 & 16 & 32 & 64 & 128 & 3 & ReLU & True & 39.83 & 90.21 \\[2pt]

2 & 0.001 & 16 & 32 & 64 & 128 & 3 & Tanh & True & 36.46 & 86.52 \\[2pt]

3 & 0.001 & 16 & 32 & 64 & 128 & 3 & SiLU & True & 41.75 & 86.53 \\[2pt]

\vdots & \vdots & \vdots & \vdots & \vdots & \vdots & \vdots & \vdots & \vdots & \vdots & \vdots \\[2pt]

384 & 0.0001 & 64 & 64 & 128 & 256 & 5 & SiLU & False & 37.44 & 84.65 \\
\midrule
&   &   &  &  &  &  &   &\multicolumn{2}{c}{\sc{total duration}} & $\approx$ 7.6-hr \\
\bottomrule
\end{tabular}
\end{small}
\vspace{-0.08in}
\end{table*}
\begin{table*}[ht]
\caption{Sample entries from the DNN coarse-grained search lookup table.}
\label{tab:dnn_lookup_sample}
\centering
\begin{small}
\begin{tabular}{c c c c c c c c c c}
\toprule
\multirow{2}{*}{Run} &

$p_1$ &
$p_2$ &
$p_3$ &
$p_4$ &
$p_5$ &
$p_6$ &
$p_7$ &
$y_1$ &
$t$ \\[2pt]

 &
($\mathrm{lr}$) &
(Batch) &
($n_\mathrm{layers}$) &
($n_\mathrm{neurons}$) &
(Activation) &
(Batch Norm) &
(Dropout) &
(Accuracy, \%) &
(sec) \\
\midrule
1 & 0.001 & 32 & 2 & 64  & ReLU & True  & 0.0 & 84.75 & 200.33 \\[2pt]

2 & 0.001 & 32 & 2 & 64  & ReLU & False & 0.0 & 83.09 & 119.01 \\[2pt]

3 & 0.001 & 32 & 2 & 128 & ReLU & True  & 0.0 & 84.19 & 203.54 \\[2pt]

\vdots & \vdots & \vdots & \vdots & \vdots & \vdots & \vdots & \vdots & \vdots & \vdots \\[2pt]

288 & 0.0001 & 128 & 4 & 128  & SiLU & False  & 0.3 & 85.46 & 26.43 \\
\midrule
&   &  &  &  &  &   & \multicolumn{2}{c}{\sc{total duration}} & $\approx$ 6.3-hr \\
\bottomrule
\end{tabular}
\end{small}
\vspace{-0.08in}
\end{table*}

\end{document}